\documentclass[letterpaper, 10 pt, conference]{ieeeconf}  

\usepackage[utf8]{inputenc}
\usepackage[usenames, dvipsnames]{color}
\usepackage{graphicx}
\usepackage{enumerate}
\usepackage{multirow}
\usepackage{graphics}
\usepackage{svg}
\usepackage{graphicx}

\usepackage{amsmath}
\usepackage{nth}
\usepackage{makecell}
\usepackage{colortbl}

\usepackage{tikz}
\usetikzlibrary{shapes,arrows}
\usepackage{subcaption}
\usepackage{amsmath,amsthm,amssymb,mathrsfs} 
\usepackage[final]{pdfpages}
\usepackage[]{placeins,flafter}
\usepackage[none]{hyphenat} 
\usepackage{xcolor}
\usepackage{adjustbox}
\usepackage{balance}
\usepackage{comment}
\usepackage{nicefrac}
\usepackage{url}
\usepackage{hyperref}

\usepackage{calrsfs}
\hypersetup{hidelinks}

\DeclareMathAlphabet{\pazocal}{OMS}{zplm}{m}{n}
\newcommand{\Lb}{\pazocal{L}}

\newcommand{\Sb}{\pazocal{S}}

\newcommand\blfootnote[1]{%
  \begingroup
  \renewcommand\thefootnote{}\footnote{#1}%
  \addtocounter{footnote}{-1}%
  \endgroup
}

\DeclareMathOperator*{\argmin}{arg\,min}

\IEEEoverridecommandlockouts                              

\title{\LARGE \bf
Endo-Depth-and-Motion: Reconstruction and Tracking in Endoscopic Videos using Depth Networks and Photometric Constraints
\vspace{-4mm}}

\author{David Recasens, Jos{\'e} Lamarca, Jos{\'e} M. F{\'a}cil, J. M. M. Montiel, Javier Civera
\thanks{
        {\tt\small \{recasens,jlamarca,jmfacil,josemari,jcivera\} @unizar.es}}
}

\begin{document}
\twocolumn[{%
\renewcommand\twocolumn[1][]{#1}%
\maketitle
\begin{center}
    \centering
    \includegraphics[width=.97\textwidth,keepaspectratio]{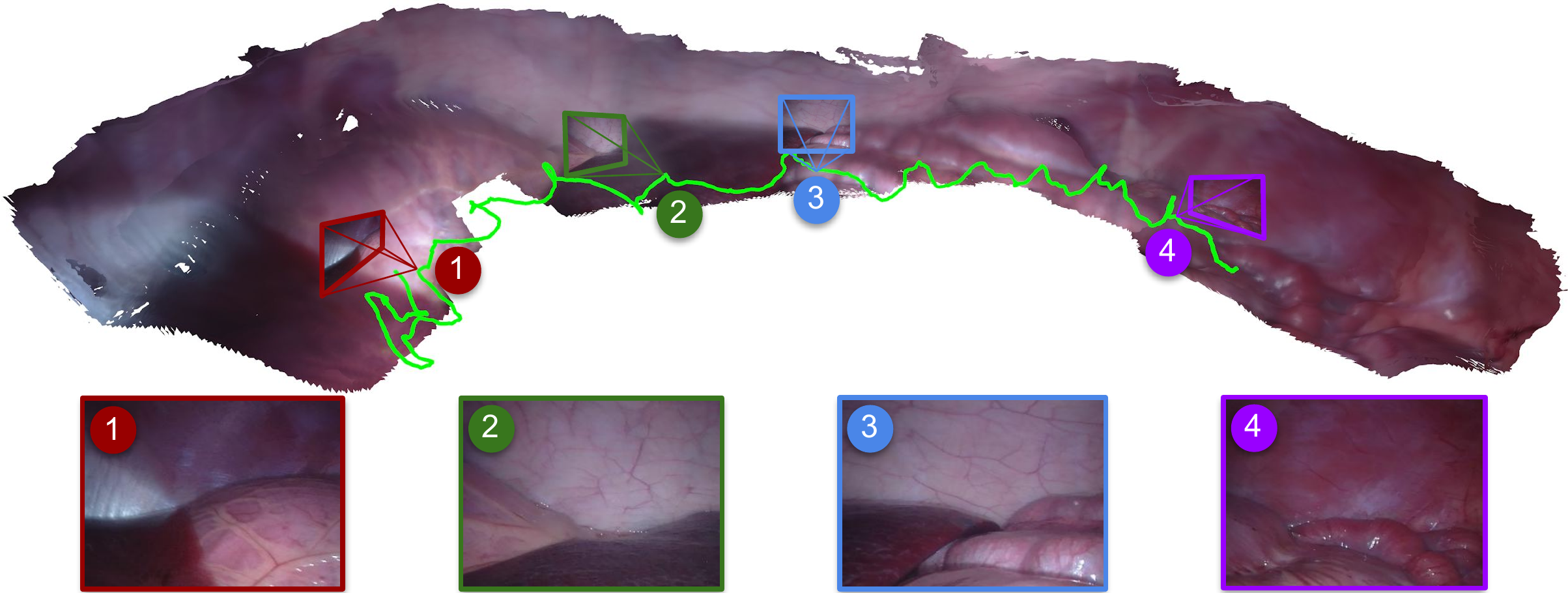}
    \captionof{figure}{\centering Top: abdominal cavity reconstruction and camera trajectory (green line) estimation by \emph{Endo-Depth-and-Motion} from a monocular sequence from video \#22 of the {Hamlyn} dataset. Bottom: four sample images from the video.\vspace{10pt}}
    \label{fig:teaser_image}
\end{center}
}]

\thispagestyle{empty}
\pagestyle{empty}


\begin{abstract}
Estimating a scene reconstruction and the camera motion from in-body videos is challenging due to several factors, \emph{e.g.} the deformation of in-body cavities or the lack of texture. In this paper we present \emph{Endo-Depth-and-Motion}, a pipeline that estimates the 6-degrees-of-freedom camera pose and dense 3D scene models from monocular endoscopic sequences. Our approach leverages recent advances in self-supervised depth networks to generate \emph{pseudo-}RGBD frames, then tracks the camera pose using photometric residuals and fuses the registered depth maps in a volumetric representation. We present an extensive experimental evaluation in the public dataset {Hamlyn}, showing high-quality results and comparisons against relevant baselines. We also release all models and code\footnote{
\url{https://davidrecasens.github.io/EndoDepthAndMotion/}} for future comparisons.\blfootnote{\scriptsize This work was supported by EndoMapper GA 863146 (EU-H2020), PGC2018-096367-B-I00 (Spanish Government), DGA-T45\_17R/FSE (Arag\'on Government). The authors are with I3A, Universidad de Zaragoza, Spain.
        {\tt\scriptsize \{recasens,jlamarca,jmfacil,josemari,jcivera\}@unizar.es}}
\hspace{-5mm}
\end{abstract}


\section{Introduction}
\label{sec:1}

Estimating a 3D reconstruction of a scene from a set of images, together with the poses of the cameras that captured them, is most of the times thought of as a mature technology. The scientific literature has shown impressive results in a wide variety of settings: we can differentiate for example between offline~\cite{schonberger2016structure} and online~\cite{mur2017orb} approaches; and among the online ones, feature-based~\cite{mur2017orb}, semi-dense{~\cite{engel2014lsd}}, hybrid~\cite{lee2018loosely} or fully dense ones~\cite{czarnowski2020deepfactors}. Such impressive results, however, rely on several assumptions frequently overlooked, namely a sufficiently rigid and textured scene, sufficient and stable illumination, and a sufficiently large camera translation --but not so large that finding correspondences becomes problematic. The application of state-of-the-art 3D vision algorithms becomes challenging in certain real-world applications where these assumptions do not hold.

A particularly challenging case are medical images, and our specific application of monocular recordings from endoscopic procedures. See Fig.~\ref{fig:teaser_image} and notice the insufficient and unstable illumination or the lack of texture. However, it is of high relevance and interest creating 3D models of the human cavities and localizing a camera inside it to enable, among others, {virtual augmentations in surgical procedures \cite{mahmoud2017patient}, assistance in polyp detection~\cite{van2006polyp,ma2019real}} and in-body autonomous robot navigation \cite{fagogenis2019autonomous}. In this paper we target dense reconstructions of in-body cavities and accurate ego-motion from monocular endoscopic sequences. Specifically, we leverage depth convolutional networks to create \emph{pseudo-}RGBD keyframes, we estimate the camera motion using photometric methods and fuse the registered \emph{pseudo-}RGBD keyframes in a volumetric representation. We achieve high-quality reconstructions in the {Hamlyn} dataset (see an example in Fig.~\ref{fig:teaser_image}). Our pipeline, that we call \emph{Endo-Depth-and-Motion}, is {among the first ones} based on depth convolutional networks that produces accurate and dense 3D reconstructions of in-body cavities. We show results that are competitive against relevant baselines (IsoNRSfM~\cite{parashar2017IsoNRSfM} {and LapDepth \cite{song2021monocular}}) and provide an extensive evaluation of the pipeline. 
\section{Related work}
\label{sec:2}

\begin{figure*}[ht!]
    \centering
    \includegraphics[width=.87\textwidth,keepaspectratio]{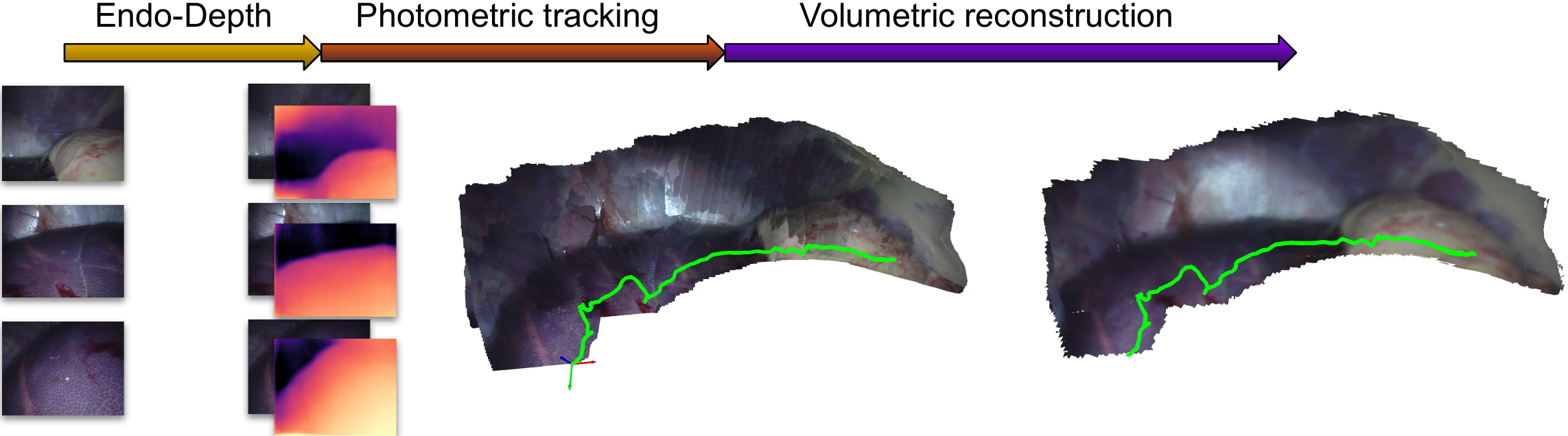}
    \captionof{figure}{
    \emph{Endo-Depth-and-Motion} overview. First, monocular depth is predicted with \emph{Endo-Depth}. The camera motion (green line) is then estimated using the photometric residuals. Finally, the full scene is reconstructed by volumetric fusion. 
    }
    \label{fig:overview}
\end{figure*}




Deep convolutional networks were first proposed for depth estimation in \cite{eigen2014depth, eigen2015predicting}. These first models were trained in a supervised manner using depth sensors, and were improved in following works using deeper networks and several architectural contributions \cite{laina2016deeper,fu2018deep,facil2019cam, song2021monocular}. Self-supervised approaches were proposed later, mainly based on multi-view photometric consistency. \cite{godard2017unsupervised} proposed self-supervision using stereo, and \cite{zhou2017unsupervised} generalized it to monocular views (ego-motion being predicted by another network). Both have been improved in many recent works, e.g., \cite{zhan2018unsupervised,godard2019digging}. {Self-supervised depth learning is relevant for in-body monocular reconstructions, as it does not require additional sensors, sophisticated renderings~\cite{mahmood2018deep,incetan2021vr} or domain transfer~\cite{mahmood2018unsupervised}. In medical contexts, supervised depth learning has been addressed in \cite{visentini2017deep,shen2019context,rau2019implicit}, self-supervised using stereo in \cite{xu2019unsupervised} and self-supervised using sparse monocular depth in \cite{liu2019dense}. \cite{chen2019slam} is the first that fuses the monocular depths from a supervised network using ElasticFusion~\cite{whelan2016elasticfusion}. Differently from them, in \emph{Endo-Depth} we use the state-of-the-art network of \cite{godard2019digging} with the following advantages: 1) the training is self-supervised from monocular endoscopes, stereo ones, or from both, 2) our experiments show that self-supervised training removes the effect of the domain shift from synthetic data, 3) its dense photometric loss uses the information of the whole image, and 4) several details (e.g, minimum reprojection loss from a set of images) make the loss sufficiently robust to endoscopic image challenges.}


Depth and motion can be estimated using only multi-view constraints, either dense~\cite{newcombe2011dtam} or sparse~\cite{mur2017orb}. The literature on these methods is huge, but they are limited on in-body images by drastic illumination changes, weak texture, deformations, tool insertions, fluids and sometimes small camera motions. All of this poses challenges in determining the correspondences and the geometry, that can be alleviated starting from single-view dense depth as we do in \emph{Endo-Depth-and-Motion}. Convolutional networks are indeed demonstrating a huge potential to replace several parts of SfM/SLAM pipelines~\cite{cadena2016past}. 
\cite{facil2017single,tateno2017cnn} are early works combining single-view depth with multi-view approaches for mapping and tracking respectively. \cite{zhou2020deeptam} proposed a convolutional model for camera tracking and incremental mapping, in this case tightly integrating multi-view optimization within the network. \cite{czarnowski2020deepfactors} uses variational auto-encoders for single-view depth and optimizes the depth prediction jointly with the camera poses. The combination we propose is photometric odometry, using the self-supervised \emph{Endo-Depth} prediction as a \emph{pseudo-}RGBD keyframe.



The interest for applying SfM/SLAM in intracorporeal sequences has risen following the advances of the field, but encounters the challenges mentioned before. Early monocular approaches were based on the Extended Kalman Filter~\cite{grasa2011ekf,grasa2013visual}; and more recent ones on non-linear optimization for tracking and mapping~\cite{mahmoud2017patient} and map densification using variational approaches~\cite{mahmoud2018live} {or multi-view stereo~\cite{marmol2019dense}}. These methods were strongly based on the rigidity assumption. MIS-SLAM {\cite{song2017dynamic,song2018mis}} was the first bringing deformable SLAM to intracorporeal images. It uses a canonical shape, as DynamicFusion \cite{newcombe2015dynamicfusion}, integrating stereo observations in a Truncated Signed Distance Function (TSDF)~\cite{curless1996volumetric} with a deformation model. It uses the rigid tracking of ORB-SLAM2 \cite{mur2017orb} to estimate the camera pose between keyframes. DefSLAM~\cite{lamarca2020defslam} was the first monocular SLAM fully addressing deformations in monocular endoscopies. SD-DefSLAM \cite{rodriguez2020sd} improves over it {incorporating an illumination-invariant Lukas-Kanade tracker}, relocalization and tool segmentation. Both of them use at their core an isometric NRSfM (IsoNRSfM) \cite{parashar2017IsoNRSfM} over a sliding window and a robust deformation tracking inspired in \cite{lamarca2018camera}. Although {IsoNRSfM} models intracorporeal deformations, it assumes that the scene is a continuous surface, which does not hold for many in-body scenes. In addition, even in \cite{rodriguez2020sd}, feature correspondence keeps being a challenge. As another drawback, deformable tracking is computationally demanding. Compared to them, our \emph{Endo-Depth} can be a fair substitute of {IsoNRSfM} for deformable SLAM. And, under the assumption of slow deformations, our high-keyframe-rate odometry allows \emph{Endo-Depth-and-Motion} to achieve long tracks in both rigid and deformable in-body sequences. 
\section{Overview}
\label{sec:3}

Fig. \ref{fig:overview} shows an overview of \emph{Endo-Depth-and-Motion}. First, pixel-wise depth is predicted on a set of keyframes of the endoscopic monocular video using a deep neural network. This part is described in Section \ref{sec:4}. The motion of each frame with respect to the closest keyframe is estimated by minimizing the photometric error, robustified using image pyramids and robust error functions (see Section \ref{sec:5} for the details). Finally, the depth maps of the keyframes are fused in a TSDF-based volumetric representation, as explained in Section \ref{sec:6}. A demonstrative video showing sample results is available as supplementary material\footnote{The video is at 
\url{https://youtu.be/G1XWIyEbvPc}}.
\section{Endo-Depth}
\label{sec:4}

\begin{figure}[t!]
    \centering
    \includegraphics[width=\columnwidth,keepaspectratio]{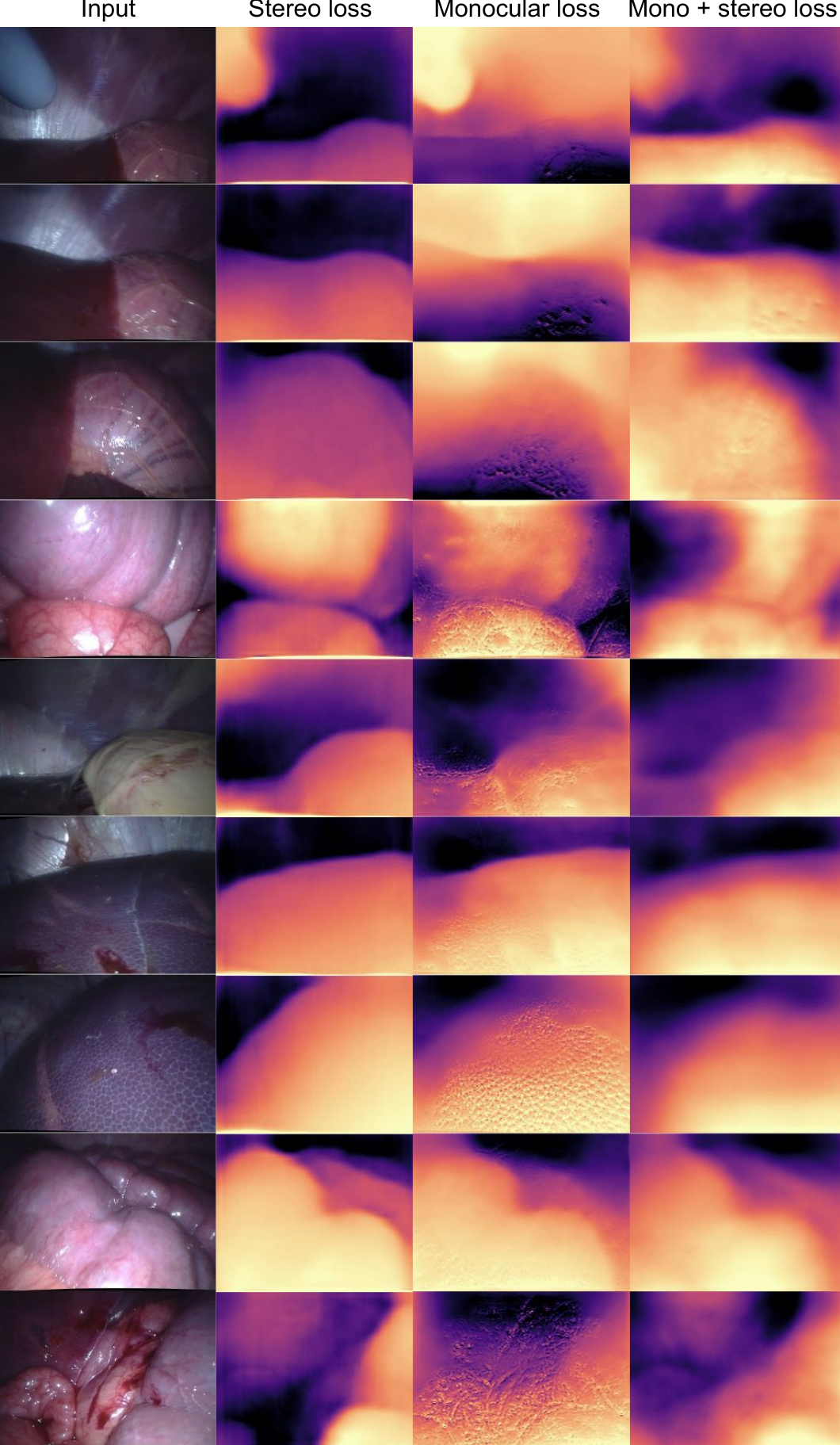}
    \captionof{figure}{
    \emph{Endo-Depth} predictions on sample images of the {Hamlyn} dataset, trained with different losses.\hspace{-25pt}}
    
    \label{fig:depths}
\end{figure}

We use the {Monodepth2} network architecture and training procedures~\cite{godard2019digging}, which are the state of the art for self-supervised depth learning. {Monodepth2} follows a U-Net encoder-decoder architecture with a ResNet18 encoder, and models the mapping function $f:\mathbb{R}^{w \times h \times 3} \rightarrow \mathbb{R}^{w \times h}$ between an image $I$ of size $w\times h$ and its depth map $D=f(I;\boldsymbol{\theta})$. Fig. \ref{fig:depths} shows several examples of predicted depth maps. The network parameters $\boldsymbol{\theta}$ are learned in a self-supervised manner minimizing a total loss $\Lb = \Lb_p + \lambda \Lb_s$, formed by  the weighted sum of a photometric loss $\Lb_p$ and a smoothing loss $\Lb_s$. This last one regularizes the surface while preserving discontinuities at the edges (see details in \cite{godard2019digging}).

{The photometric loss $\Lb_p$ sums over each pixel $\boldsymbol{p}\in \Omega_t$ in a target image $I_{t}$ the minimum value of an appearance residual with a pixel $\boldsymbol{p} \in \Omega_{t^\prime \rightarrow t}$ in images $I_{t^\prime \rightarrow t}$ that are warped from a small set of source images $\Sb=\{  I_1, \hdots, I_{t^\prime}, \hdots \}$. The network can be trained using stereo pairs --the target and source images are the stereo ones--, monocular views --the source images are the images before and after the target one--, or both. Specifically, the loss is a function of the photometric reprojection error and the Structural Similarity Index Measure (SSIM)~\cite{wang2004image}}

\begin{multline}
{
    \Lb_p = \sum_{\boldsymbol{p}\in \Omega_t} \min_{\Sb} \left( \alpha \lVert I_t\left[\boldsymbol{p}\right] - I_{t^\prime \rightarrow t}\left[\boldsymbol{p}\right] \rVert_1 + \right.} \\
    {\left. + \left(1- \text{SSIM}\left(I_t,I_{t^\prime \rightarrow t},\boldsymbol{p}\right) \right) \right)~,}
\end{multline}

{\noindent where $\left[\cdot \right]$ is the sampling operator and $\alpha$ weights the two addends. The color values of $\boldsymbol{p} \in \Omega_{t^\prime \rightarrow t}$ are taken from the corresponding pixels $\boldsymbol{p}^{\prime} \in \Omega_{t^\prime}$, so that $I_{t^\prime \rightarrow t}\left[\boldsymbol{p}\right] = I_{t^\prime}\left[\boldsymbol{p}^{\prime}\right]$. The warping function is}



\begin{equation}
   {\boldsymbol{p}^{\prime} = \pi \left( \mathbf{R}_{t^\prime t} \pi^{-1}\left(\boldsymbol{p},D_t\left[\boldsymbol{p}\right] \right) +  \mathbf{t}_{t^\prime t}\right)}
\label{eq:backandforthprojection}
\end{equation}

{\noindent where $\pi:\mathbb{R}^3 \rightarrow \Omega$ and $\pi^{-1}:\Omega \times \mathbb{R} \rightarrow \mathbb{R}^3$ are respectively the projection and back-projection functions. $\mathbf{T}_{t^\prime t} = \big(\begin{smallmatrix}
  \mathbf{R}_{t^\prime t} & \mathbf{t}_{t^\prime t}\\
  \mathbf{0} & 1
\end{smallmatrix}\big) \in \ensuremath{\mathrm{SE}(3)}$, with $\mathbf{R}_{t^\prime t} \in \ensuremath{\mathrm{SO}(3)}$ and $\mathbf{t}_{t^\prime t} \in \mathbb{R}^3$, is the transformation matrix that converts points from the target reference frame of $I_{t}$ to the source frame of $I_{t^\prime}$. Such transformation is pre-calibrated when training with a stereo pair and learned by another deep network when training with two monocular views.}



\section{Photometric Tracking}
\label{sec:5}

We use a keyframe-based photometric approach for tracking the camera pose, that estimates a relative transformation matrix $\mathbf{T}_{ck} \in \ensuremath{\mathrm{SE}(3)}$ between a target keyframe $I_k$ and the current source frame $I_c$. Given the dense depth map for the last keyframe $D_k=f(I_k;\boldsymbol{\theta})$, predicted using \emph{Endo-Depth} (Section \ref{sec:4}), we optimize $\mathbf{T}_{ck}$ so that corresponding pixels have minimum color difference.
We formulate it as a non-linear least-squares in a Lucas-Kanade style, minimizing the photometric error. To improve the range of convergence, known to be a challenge in photometric methods, we use a coarse-to-fine pyramidal optimization.

Our approach is similar to DTAM \cite{newcombe2011dtam}. First, we optimize only the camera rotation using the lowest level of the the pyramid. This gives us resilience under motion blur and helps convergence, even when the motion is not a pure rotation. For every scale after the coarser one, we optimize the full six-degrees-of-freedom camera pose. 
We parametrize local motion updates by using Lie algebras as $\boldsymbol{\psi} \in \ensuremath{\mathfrak{se}(3)}$. {The final cost function is the forward-compositional photometric cost between corresponding pixels in the keyframe $\boldsymbol{p} \in \Omega_k$ and in the warped frame $\boldsymbol{p} \in \Omega_{c \rightarrow k}$}


\begin{equation}
{
    \\ \hat{\boldsymbol{\psi}} = \argmin_{\boldsymbol{\psi}} \sum_{\boldsymbol{p}\in \Omega_k} \min \left( \lVert I_k  \left[\boldsymbol{p}\right] -I_{c \rightarrow k}\left[\boldsymbol{p}\right] \rVert_{2}, \gamma \right)
    }
\end{equation}

\noindent {where the color value of $\boldsymbol{p} \in \Omega_{c \rightarrow k}$ is taken from its correspondences $\boldsymbol{p}^{\prime} \in \Omega_c$ using unprojection and projection as in Equation \ref{eq:backandforthprojection}. The frame-to-keyframe motion $\mathbf{T}_{ck}$ is the composition of a global transformation $\mathbf{T}_{ck}^0$ and the exponential mapping of the local updates in the Lie algebra $\mathbf{T}_{ck} = \mathrm{exp}_\mathrm{SE(3)}(\boldsymbol{\psi})\mathbf{T}_{ck}^0$.} We use Gauss-Newton, which converges in a few steps. Our tracking does not model occlusions nor illumination changes, as both have a small effect in our narrow baseline setup. In any case, to mitigate the effect of outliers, we saturate the L2-norm of the residuals with a threshold $\gamma$ determined experimentally. 

\section{Volumetric Reconstruction}
\label{sec:6}

In this final stage of our pipeline, we fuse the registered \emph{pseudo-}RGBD keyframes obtained from \emph{Endo-Depth} after tracking into an implicit surface representation. Specifically, we use a TSDF~\cite{curless1996volumetric}. The scene is first divided into voxels and, for each of them it is stored a cumulative signed distance function $d:\mathbb{R}^{3} \rightarrow \mathbb{R}$ that represents the distance to the closest surface (which it is truncated at a certain depth value). The TSDF can be updated in a straightforward manner, using sequential averaging for every voxel and the predicted depth for every pixel $\boldsymbol{p} \in \Omega_k$ in every keyframe $k$. The surface can be efficiently recovered from such implicit representation using, for instance, the Marching Cubes algorithm~\cite{lorensen1987marching}. In our experiments, we use the implementation in Open3D \cite{Zhou2018}.

\begin{figure}[ht!]
    \centering
    \includegraphics[width=.39\textwidth,keepaspectratio]{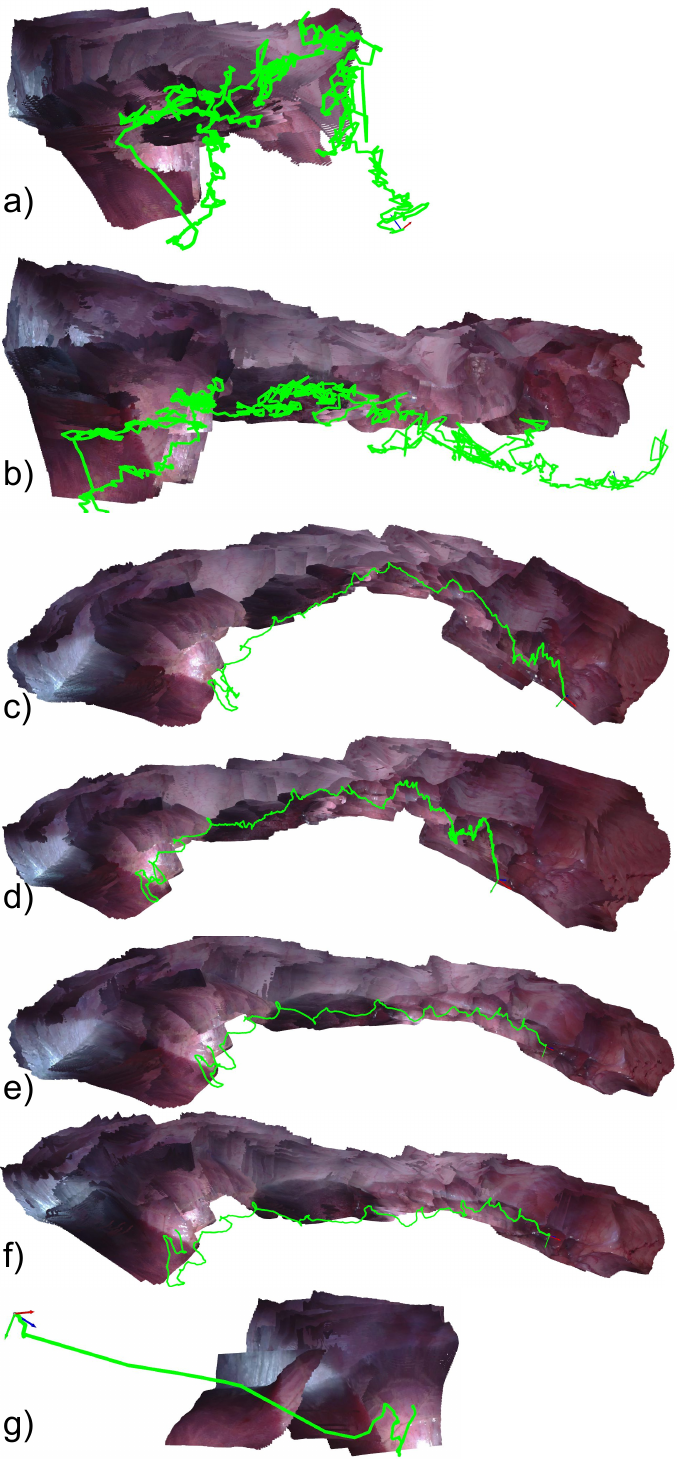}
    \captionof{figure}{Trajectory and reconstructions before fusion in {Hamlyn} video \#22 for 7 tracking alternatives: a) point-to-point ICP \cite{besl1992method}, b) point-to-plane ICP \cite{chen1992object}, c) photometric~\cite{steinbrucker2011real}, d) photometric plus geometric \cite{park2017colored} and ours with keyframe creation ratio e) $0.5$, f) $0.2$ and g) $0.1$.}
    \label{fig:odometry_comparative}
\end{figure}

\begin{figure}[ht!]
    \centering
    \includegraphics[width=.4\textwidth,keepaspectratio]{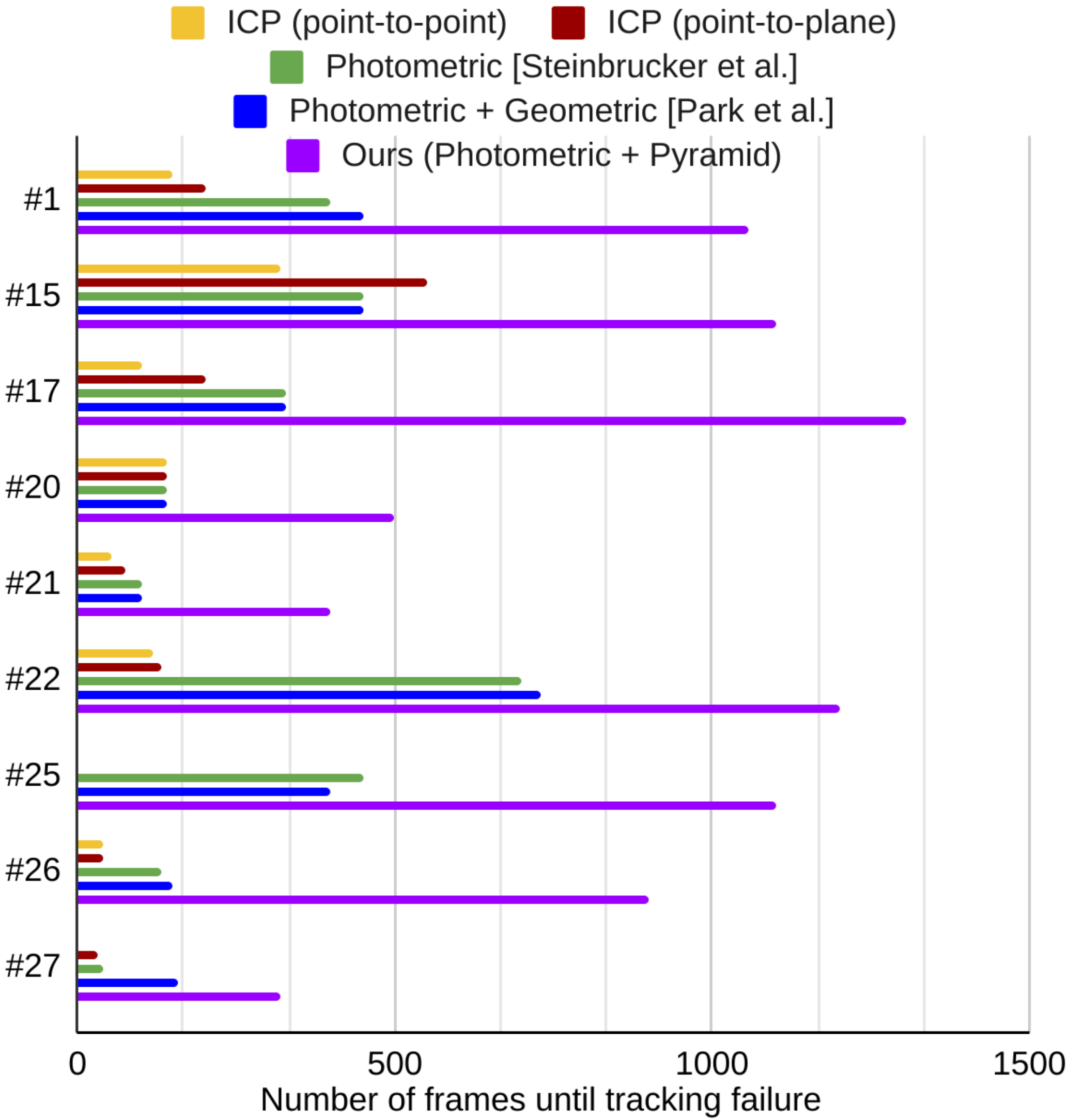}
    \captionof{figure}{Frames until failure for 5 tracking alternatives (ICP point-to-point \cite{besl1992method}, ICP point-to-plane \cite{chen1992object}, photometric \cite{steinbrucker2011real}, hybrid photometric and geometric \cite{park2017colored} and ours) in 9 {Hamlyn} videos (same as the maps in Fig. \ref{fig:teaser_image}, \ref{fig:overview} and \ref{fig:odometry_comparative}).}
    \label{fig:odometry_comparative_frames}
\end{figure}
\section{Experimental Results}
\label{sec:8}


{We used the Hamlyn dataset~\cite{mountney2010three,stoyanov2005soft,stoyanov2010real,pratt2010dynamic}, that contains challenging sequences imaging intracorporeal scenes with weak textures, deformations, reflections, surgical tools and occlusions. Specifically, we chose 21 videos with diverse image resolution and calibration parameters.}

\subsection{Endo-Depth}

Fig. \ref{fig:depths} shows \emph{Endo-Depth} predictions for several sample images of the {Hamlyn} dataset. To obtain quantitative results, we generated ground truth depth from stereo, specifically using Libelas ({Library for Efficient Large-Scale Stereo Matching}~\cite{geiger2010efficient}). It must be remarked that Libelas is sufficiently good but not perfect: it leaves blank areas for pixels at image borders and sometimes within the image. 
This ground truth is made available to be used as benchmark and can be found in our project page.

We defined the configuration for self-supervised training after an extensive ablation study. First, we set the input size to half of the original images. In this manner we filter details like veins, small folds or specular reflections that cause color changes unrelated to depth. Also, using reflection padding (instead of the more usual zero padding) in the decoder demonstrated a better performance both qualitatively and quantitatively. Many works \cite{godard2019digging,girshick2014rich,kuznietsov2017semi,guo2018learning} used an encoder pretrained for ImageNet classification \cite{russakovsky2015imagenet}, achieving smaller errors compared to training from scratch. For {Monodepth2}, pretraining is also benefitial when evaluated in  {KITTI}~\cite{geiger2013vision}, and we have seen the same effect for \emph{Endo-Depth} evaluated in {Hamlyn}. Summing up, all our models were pre-trained in ImageNet, and trained in Hamlyn halving the resolution of the input images and with reflection padding. {With stereo and monocular-stereo data, the best performance was reached after 2 epochs. When training only with monocular images, 9 epochs were necessary.}

We do 21-fold cross validation for each training modality in Tables \ref{tab:endodepth_eval_red_without_scale} and \ref{tab:endodepth_eval_scale_red}. Each of the 21 models is tested on one {Hamlyn} video and trained on the rest. 
To address the varying camera intrinsics in the {Hamlyn} data, we grouped together videos that have similar resolution and intrinsics and trained \emph{Endo-Depth} sequentially in each of them. For the training splits we removed occluded or useless frames, and for the test ones, also those frames with low-quality ground truth. We report the standard metrics for depth networks~\cite{eigen2014depth}. As monocular depth is affected of scale drift, in particular if trained with monocular sequences, we align the scales of the predicted depth $D_i$ and the ground truth depth $D_i^{GT}$ using a per-image scale factor $s_i$ computed, as it is standard in the literature{~\cite{eigen2014depth}}, as $s_i = \nicefrac{\text{median}(D_i^{GT})}{\text{median}(D_i)}$. We show in the last column of Table \ref{tab:endodepth_eval_red_without_scale} that the stereo loss achieves the best performance in most of the sequences. This is expected, since the illumination does not change nor the scene deforms between the images of a stereo pair. On the other side, the monocular loss is affected by these two factors, in addition to having to learn the relative camera motion.




Table \ref{tab:endodepth_eval_scale_red} shows \emph{Endo-Depth} results {\bf without} depth scaling, as stereo self-supervision observes such scale. We consider this a more realistic setup and a more relevant metric for applications. Notice that the differences with per-image scaling are small (less than $1$ cm RMSE in average). This demonstrates that \emph{Endo-Depth} trained with monocular-stereo or stereo losses learns a stable global scale for the reconstruction, which is a key aspect to obtain small trajectory drift in the experiments of Section \ref{sec:camtracking}.

\begin{table*}[t!]
    \normalsize
    \centering
    \resizebox{\textwidth}{!} {%
    \begin{tabular}{ c  c  c  c  c  c  c  c  c  c  c  c  c  c  c  c  c  c  c  c  c  c  c  r  }
        \multicolumn{2}{c}{} & \multicolumn{22}{c}{\textbf{Sequence}} \\
        \noalign{\vskip 1mm} 
        \cline{3-23}
        \noalign{\vskip 1mm}   
        \multicolumn{1}{c}{} & \multicolumn{1}{c}{\textbf{Loss}} & 1 & 4 & 5 & 6 & 8 & 11 & 12 & 14 & 15 & 16 & 17 & 18 & 19 & 20 & 21 & 22 & 23 & 24 & 25 & 26 & \multicolumn{1}{c}{27} & \textbf{\#-Best} \\
        \noalign{\vskip 1mm} 
        \hline
        \noalign{\vskip 1mm}      
        \cellcolor{Dandelion} & Mono  &   0.260  &   0.059  &   0.064  &   0.152  &   0.132  &   0.112  &   0.089  &   0.162  &   0.109  &   0.100  &   0.232  &   0.308  &   0.296  &   0.115  &   0.123  &   0.230  &   0.252  &   0.241  &   0.146  &   0.157  &   0.178   &   0/21 \\ 
        \cellcolor{Dandelion} & Stereo   &   \textbf{0.091}  &   \textbf{0.038}  &   \textbf{0.047}  &   \textbf{0.080}  &   0.120  &   0.092  &   \textbf{0.040}  &   \textbf{0.136}  &   \textbf{0.070}  &   \textbf{0.072}  &   \textbf{0.195}  &   \textbf{0.230}  &   \textbf{0.184}  &   0.109  &   \textbf{0.108}  &   \textbf{0.158}  &   \textbf{0.166}  &   \textbf{0.183}  &   \textbf{0.141}  &   \textbf{0.131}  &   \textbf{0.128}  &   \textbf{18/21}  \\ 
        \cellcolor{Dandelion} \multirow{-3}{*}{Abs Rel}& Mono + Stereo   &   0.193  &   0.040  &   0.073  &   0.432  &   \textbf{0.114}  &   \textbf{0.079}  &   0.109  &   0.163  &   0.088  &   0.095  &   0.222  &   0.232 &   0.272  &   \textbf{0.095}  &   0.137  &   0.211  &   0.215  &   0.201  &   0.171  &   0.168  &   0.152   &   3/21\\ 
        \noalign{\vskip 1mm} 
        \hline
        \noalign{\vskip 1mm}     
        \cellcolor{Dandelion} & Mono  &  19.775  &   0.269  &   0.335  &   6.461  &   2.738  &   1.491  &   1.629  &   3.859  &   0.796  &   1.102  &   6.175  &  10.802  &  12.437  &   1.093  &   1.681  &   5.689  &   9.159  &   6.260  &   \textbf{2.318}  &   2.874  &   4.247   &   1/21 \\ 
        \cellcolor{Dandelion}  & Stereo    &   \textbf{1.429}  &   \textbf{0.160}  &   \textbf{0.186}  &   \textbf{1.057}  &   1.951  &   1.332  &   \textbf{0.651}  &   \textbf{2.855}  &   \textbf{0.425}  &   \textbf{0.751}  &   \textbf{5.344}  &   6.994  &   \textbf{4.379}  &   1.084  &   \textbf{1.581}  &   \textbf{3.147}  &   \textbf{4.954}  &   \textbf{3.664}  &   2.471  &   \textbf{2.407}  &   \textbf{2.371}   &   \textbf{16/21} \\ 
        \cellcolor{Dandelion} \multirow{-3}{*}{Sq Rel}&Mono +  Stereo  &   9.251  &   0.193  &   0.458  &  56.218  &   \textbf{1.906}  &   \textbf{0.931}  &   1.442  &   3.861  &   0.643  &   1.111  &   5.676  &   \textbf{6.415}  &   9.699  &   \textbf{0.723}  &   2.344  &   4.772  &   7.936  &   4.136  &   3.224  &   3.072  &   3.204   &   4/21 \\
        \noalign{\vskip 1mm} 
        \hline
        \noalign{\vskip 1mm}   
        \cellcolor{Dandelion} & Mono  &  30.617  &   3.405  &   3.859  &  23.488  &  14.301  &   9.978  &  10.232  &  14.142  &   5.279  &   6.796  &  19.540  &  28.302  &  24.495  &   7.057  &   9.272  &  18.805  &  21.619  &  18.321  &  \textbf{10.754}  &  12.872  &  15.517    &   1/21 \\ 
        \cellcolor{Dandelion} &  Stereo   &   \textbf{9.792}  &   \textbf{2.385}  &   \textbf{2.855}  &  \textbf{10.653}  &  \textbf{11.296}  &   9.156  &   \textbf{6.156}  &  \textbf{11.833}  &   \textbf{3.537}  &   \textbf{5.258 } &  \textbf{17.793}  &  \textbf{21.772}  &  \textbf{15.723}  &   6.816  &   \textbf{8.760}  &  \textbf{13.848}  &  \textbf{16.003}  &  \textbf{14.522}  &  11.022  &  \textbf{11.862}  &  \textbf{11.072}   &   \textbf{18/21} \\ 
        \cellcolor{Dandelion}\multirow{-3}{*}{RMSE} &Mono +  Stereo  &  21.042  &   2.572  &   4.472  &  55.892  &  11.729  &  \textbf{ 7.790 } &  10.160  &  13.068  &   4.798  &   6.648  &  18.868  &  21.984  &  22.274  &   \textbf{5.735}  &  10.657  &  17.145  &  19.263  &  15.523  &  12.489  &  13.506  &  12.570   &   2/21 \\ 
        \noalign{\vskip 1mm} 
        \hline
        \noalign{\vskip 1mm}   
        \cellcolor{Dandelion}& Mono  &   0.287  &   0.075  &   0.083  &   0.187  &   0.188  &   0.159  &   0.126  &   0.217  &   0.156  &   0.130  &   0.272  &   0.361  &   0.320  &   0.145  &   0.152  &   0.271  &   0.284  &   0.294  &   0.186  &   0.197  &   0.213   &   0/21 \\ 
        \cellcolor{Dandelion} & Stereo   &   \textbf{0.114}  &   \textbf{0.052}  &   \textbf{0.061}  &   \textbf{0.099}  &   0.154  &   0.126  &   \textbf{0.070}  &   \textbf{0.183}  &   \textbf{0.099}  &   \textbf{0.099}  &   \textbf{0.239}  &   0.279  &   \textbf{0.230}  &   0.138  &   \textbf{0.141}  &   \textbf{0.199}  &   \textbf{0.206}  &   \textbf{0.226}  &   \textbf{0.179}  &   \textbf{0.169}  &   \textbf{0.161}   &   \textbf{17/21} \\ 
        \cellcolor{Dandelion} \multirow{-3}{*}{RMSE\textsubscript{log}} &Mono +  Stereo  &   0.230  &   0.056  &   0.093  &   0.400  &   \textbf{0.153}  &   \textbf{0.112}  &   0.146  &   0.212  &   0.144  &   0.123  &   0.262  &    \textbf{0.278}  &   0.310  &   \textbf{0.118}  &   0.173  &   0.251  &   0.254  &   0.245  &   0.213  &   0.208  &   0.185   &   4/21\\
        \noalign{\vskip 1mm} 
        \hline
        \noalign{\vskip 1mm}   
        \cellcolor{Orchid} &  Mono  &   0.747  &   \textbf{0.989}  &   0.995  &   0.810  &   0.810  &   0.864  &   0.921  &   0.767  &   0.887  &   0.915  &   0.586  &   0.453  &   0.557  &   0.865  &   0.860  &   0.567  &   0.647  &   0.595  &   0.804  &   0.776  &   0.735   &   1/21 \\ 
        \cellcolor{Orchid} &  Stereo   &   \textbf{0.928}  &   \textbf{0.989}  &   \textbf{0.999}  &   \textbf{0.952}  &   \textbf{0.866}  &   0.941  &   \textbf{0.984}  &   \textbf{0.808}  &   \textbf{0.952}  &   \textbf{0.948}  &   \textbf{0.717}  &   \textbf{0.628}  &   \textbf{0.727}  &   0.876  &   \textbf{0.885}  &   \textbf{0.761}  &   \textbf{0.810}  &   \textbf{0.711}  &   \textbf{0.832}  &   \textbf{0.837}  &   \textbf{0.842}   &   \textbf{19/21} \\ 
        \cellcolor{Orchid}\multirow{-3}{*}{$\delta < 1.25^1$} & Mono +  Stereo  &   0.780  &   0.987  &   0.968  &   0.610  &   0.857  &   \textbf{0.962}  &   0.888  &   0.776  &   0.908  &   0.917  &   0.630  &   0.574  &   0.583  &   \textbf{0.930}  &   0.838  &   0.619  &   0.754  &   0.642  &   0.768  &   0.738  &   0.791   &   2/21 \\
        \noalign{\vskip 1mm} 
        \hline
        \noalign{\vskip 1mm}   
        \cellcolor{Orchid} &  Mono  &   0.890  &   \textbf{1.000}  &   \textbf{1.000}  &   0.963  &   0.955  &   0.989  &   0.984  &   0.924  &   0.975  &   0.988  &   0.884  &   0.764  &   0.823  &   0.993  &   \textbf{0.985}  &   0.891  &   0.856  &   0.872  &   \textbf{0.962}  &   0.956  &   0.942   &   4/21 \\ 
        \cellcolor{Orchid}&  Stereo   &   \textbf{0.989}  &   0.999  &   \textbf{1.000}  &   \textbf{0.999}  &   \textbf{0.980}  &   0.989  &   0.991  &   \textbf{0.950}  &   \textbf{0.992}  &   \textbf{0.990}  &   \textbf{0.915}  &   0.870  &   \textbf{0.921}  &   0.990  &   0.980  &   \textbf{0.955}  &   \textbf{0.923}  &   \textbf{0.930}  &   0.960  &   \textbf{0.967}  &   \textbf{0.972}   &   \textbf{14/21} \\ 
        \cellcolor{Orchid} \multirow{-3}{*}{$\delta < 1.25^2$}& Mono +  Stereo  &   0.946  &   0.999  &   0.999  &   0.861  &   \textbf{0.980}  &   \textbf{0.993}  &   \textbf{0.993}  &   0.936  &   0.971  &   0.985  &   0.893  &   \textbf{0.873}  &   0.842  &   \textbf{0.999}  &   0.962  &   0.910  &   0.881  &   0.922  &   0.944  &   0.952  &   0.959   &   5/21 \\ 
        \noalign{\vskip 1mm} 
        \hline
        \noalign{\vskip 1mm}   
        \cellcolor{Orchid} &  Mono  &   0.940  &   \textbf{1.000}  &   \textbf{1.000}  &   0.989  &   0.989  &   0.994  &   0.995  &   0.978  &   0.995  &   0.997  &   0.980  &   0.937  &   0.941  &   \textbf{1.000}  &   \textbf{0.998}  &   0.986  &   0.940  &   0.959  &   \textbf{0.995}  &   0.993  &   0.990   &   5/21 \\ 
        \cellcolor{Orchid} &  Stereo   &   \textbf{0.998}  &   \textbf{1.000}  &   \textbf{1.000}  &   \textbf{1.000}  &   0.996  &   0.995  &   0.995  &   \textbf{0.986}  &   \textbf{0.999}  &   0.998  &   0.977  &   0.967  &   \textbf{0.978}  &   0.999  &   \textbf{0.998}  &   \textbf{0.991}  &   \textbf{0.975}  &   \textbf{0.992}  &   0.993  &   \textbf{0.994}  &   \textbf{0.995}   &   \textbf{13/21} \\ 
        \cellcolor{Orchid} \multirow{-3}{*}{$\delta < 1.25^3$} & Mono +  Stereo  &   0.974  &   \textbf{1.000}  &   \textbf{1.000}  &   0.898  &   \textbf{0.997}  &   \textbf{0.996}  &   \textbf{0.996}  &   0.977  &   0.993  &   \textbf{0.999}  &   \textbf{0.985}  &   \textbf{0.983}  &   0.948  &   \textbf{1.000}  &   0.996  &   0.990  &   0.945  &   0.989  &   0.985  &   \textbf{0.994}  &   0.991   &   10/21 \\
        \noalign{\vskip 1mm} 
        \hline

    \end{tabular}%
    }
    \caption{\emph{Endo-Depth} results in {Hamlyn} with per-image scaling. $\text{Abs}_{\text{rel}}$ and RMSE in [mm], $\text{RMSE}_{\text{log}}$ in [log mm]. We compare three losses: pure monocular (Mono), stereo (Stereo) and a combination of both (Mono + Stereo). \nth{1} column: metrics used for evaluation. \nth{2} column: losses used during training. Last column: number of times a method is the best for a given metric. Rest of the columns: results for the {Hamlyn} sequences.}
    \label{tab:endodepth_eval_red_without_scale}
\end{table*}

\begin{table*}[t!]
    \normalsize
    \centering
    \resizebox{\textwidth}{!} {%
    \begin{tabular}{ c  c  c  c  c  c  c  c  c  c  c  c  c  c  c  c  c  c  c  c  c  c  c  r  }
        \multicolumn{2}{c}{} & \multicolumn{22}{c}{\textbf{Sequence}} \\
        \noalign{\vskip 1mm} 
        \cline{3-23}
        \noalign{\vskip 1mm}   
        \multicolumn{1}{c}{} & \multicolumn{1}{c}{\textbf{Loss}} & 1 & 4 & 5 & 6 & 8 & 11 & 12 & 14 & 15 & 16 & 17 & 18 & 19 & 20 & 21 & 22 & 23 & 24 & 25 & 26 & \multicolumn{1}{c}{27} & \textbf{\#-Best} \\
        \noalign{\vskip 1mm} 
        \hline
        \noalign{\vskip 1mm}      
        \cellcolor{Dandelion} & Stereo   &   0.573  &   \textbf{0.070}  &   \textbf{0.062}  &   \textbf{0.342}  &  \textbf{ 0.642}  &   0.168  &   \textbf{0.203}  &   \textbf{0.209}  &   \textbf{0.105}  &   \textbf{0.174}  &   \textbf{0.222}  &   \textbf{0.255}  &   \textbf{0.236}  &  \textbf{ 0.192}  &   \textbf{0.128}  &   \textbf{0.198}  &   \textbf{0.182}  &   \textbf{0.365}  &   \textbf{0.158}  &   \textbf{0.156}  &   0.224  & \textbf{18/21}   \\ 
        \cellcolor{Dandelion} \multirow{-2}{*}{Abs Rel} &Mono +  Stereo   &    \textbf{0.312}  &   \textbf{0.070}  &   0.079  &   0.571  &   0.754  &   \textbf{0.086}  &   0.295  &   0.514  &   0.719  &   0.396  &   0.343  &   0.262  &   0.379  &   0.279  &   0.208  &   0.267  &   0.273  &   0.447  &   0.420  &   0.347  &   \textbf{0.189}  &  4/21\\
        \noalign{\vskip 1mm} 
        \hline
        \noalign{\vskip 1mm}     
        \cellcolor{Dandelion}  & Stereo    &  29.813  &   \textbf{0.337}  &   \textbf{0.262}  &  \textbf{19.045 } &  \textbf{32.455}  &   2.724  &   \textbf{3.323}  &   \textbf{3.916}  &   \textbf{0.576}  &   \textbf{2.375}  &   \textbf{8.182}  &   8.428  &   \textbf{6.662}  &   \textbf{3.429}  &   \textbf{1.713}  &   \textbf{5.538}  &   \textbf{5.478}  &  \textbf{13.119}  &   \textbf{2.932}  &   \textbf{2.919}  &   5.067  &  \textbf{17/21}   \\
        \cellcolor{Dandelion} \multirow{-2}{*}{Sq Rel} & Mono +  Stereo   &  \textbf{18.939}  &   0.392  &   0.530  &  83.030  &  41.415  &   \textbf{1.057}  &   6.415  &  13.947  &  15.758  &   9.142  &  12.983  &   \textbf{7.743}  &  12.800  &   4.599  &   4.640  &   8.038  &   7.842  &  17.415  &  13.131  &   9.939  &   \textbf{3.928}  &  4/21\\
        \noalign{\vskip 1mm} 
        \hline
        \noalign{\vskip 1mm}   
        \cellcolor{Dandelion} &  Stereo   &  44.538  &   \textbf{3.605}  &   \textbf{3.570}  &  \textbf{29.569}  &  \textbf{38.724}  &  12.649  &  \textbf{15.521}  &  \textbf{13.724}  &   \textbf{4.152}  &   \textbf{9.971}  &  \textbf{20.678}  &  \textbf{24.102}  &  \textbf{18.435}  &  \textbf{11.130}  &   \textbf{9.605}  &  \textbf{16.963}  &  \textbf{16.622}  &  \textbf{24.723}  &  \textbf{11.799}  &  \textbf{13.138}  &  17.851  &  \textbf{18/21}   \\
        \cellcolor{Dandelion}\multirow{-2}{*}{RMSE} &Mono +  Stereo   &  \textbf{27.863}  &   3.690  &   4.799  &  65.822  &  44.831  &   \textbf{8.237}  &  20.921  &  23.659  &  20.948  &  19.805  &  26.075  &  24.948  &  30.643  &  13.813  &  14.505  &  20.749  &  21.462  &  28.698  &  24.522  &  22.421  &  \textbf{15.660}  &  3/21 \\
        \noalign{\vskip 1mm} 
        \hline
        \noalign{\vskip 1mm}   
        \cellcolor{Dandelion} & Stereo   &   0.452  &   \textbf{0.081}  &   \textbf{0.075}  &   \textbf{0.273 } &   \textbf{0.484}  &   0.183  &   \textbf{0.245}  &   \textbf{0.232}  &   \textbf{0.130}  &   \textbf{0.181}  &   \textbf{0.269}  &   \textbf{0.304}  &   \textbf{0.271}  &   \textbf{0.200}  &   \textbf{0.155}  &   \textbf{0.233}  &   \textbf{0.216}  &   \textbf{0.352}  &   \textbf{0.195}  &   \textbf{0.195}  &   0.294  &  \textbf{18/21} \\
        \cellcolor{Dandelion} \multirow{-2}{*}{RMSE\textsubscript{log}} &Mono +  Stereo  &   \textbf{0.306}  &   0.083  &   0.100  &   0.472  &   0.547  &   \textbf{0.120}  &   0.388  &   0.431  &   0.546  &   0.347  &   0.353  &   0.311  &   0.550  &   0.263  &   0.226  &   0.292  &   0.289  &   0.407  &   0.385  &   0.334  &   \textbf{0.242}  &  3/21 \\ 
        \noalign{\vskip 1mm} 
        \hline
        \noalign{\vskip 1mm}   
        \cellcolor{Orchid} &  Stereo   &   0.158  &   \textbf{0.990}  &   \textbf{0.999}  &   \textbf{0.632}  &   \textbf{0.199}  &   0.800  &   \textbf{0.488}  &   \textbf{0.696}  &  \textbf{ 0.941}  &   \textbf{0.837}  &   \textbf{0.763}  &   \textbf{0.555}  &   \textbf{0.663}  &   \textbf{0.718}  &   \textbf{0.852}  &   \textbf{0.733}  &   \textbf{0.815}  &   \textbf{0.524}  &   \textbf{0.772}  &   \textbf{0.774 } &   0.467  &  \textbf{18/21} \\
        \cellcolor{Orchid}\multirow{-2}{*}{$\delta < 1.25^1$} & Mono +  Stereo  &   \textbf{0.606}  &   0.984  &   0.961  &   0.455  &   0.153  &   \textbf{0.956}  &   0.160  &   0.299  &   0.071  &   0.276  &   0.569  &   0.486  &   0.182  &   0.470  &   0.700  &   0.613  &   0.529  &   0.413  &   0.401  &   0.438  &   \textbf{0.609}  &  3/21 \\
        \noalign{\vskip 1mm} 
        \hline
        \noalign{\vskip 1mm}   
        \cellcolor{Orchid}&  Stereo   &   0.492  &   \textbf{1.000}  &   \textbf{1.000}  &   \textbf{0.787}  &   \textbf{0.545}  &   0.983  &   \textbf{0.983}  &   \textbf{0.935}  &   \textbf{0.988}  &   \textbf{0.972}  &   \textbf{0.867}  &   \textbf{0.840}  &   \textbf{0.867}  &  \textbf{ 0.940}  &   \textbf{0.989}  &   \textbf{0.902}  &   \textbf{0.917}  &   \textbf{0.769}  &   \textbf{0.956}  &   \textbf{0.960}  &   0.828  &  \textbf{18/21} \\
        \cellcolor{Orchid} \multirow{-2}{*}{$\delta < 1.25^2$}& Mono +  Stereo   &   \textbf{0.838}  &   \textbf{1.000}  &   0.999  &   0.765  &   0.381  &   \textbf{0.992}  &   0.752  &   0.595  &   0.220  &   0.810  &   0.794  &   0.822  &   0.425  &   0.938  &   0.927  &   0.839  &   0.889  &   0.698  &   0.671  &   0.802  &   \textbf{0.911}  &  4/21 \\
        \noalign{\vskip 1mm} 
        \hline
        \noalign{\vskip 1mm}   
        \cellcolor{Orchid} &  Stereo   &   0.894  &   \textbf{1.000}  &   \textbf{1.000}  &   \textbf{0.886}  &   \textbf{0.792}  &   0.994  &   \textbf{0.990}  &   \textbf{0.980}  &   \textbf{0.997}  &   \textbf{0.996}  &   \textbf{0.943}  &   0.961  &   \textbf{0.961}  &   0.996  &   \textbf{0.998}  &   \textbf{0.975}  &   \textbf{0.969}  &   \textbf{0.908}  &   \textbf{0.990}  &   \textbf{0.993}  &   0.975  &  \textbf{16/21} \\
        \cellcolor{Orchid} \multirow{-2}{*}{$\delta < 1.25^3$} & Mono +  Stereo  &   \textbf{0.958}  &   \textbf{1.000}  &   \textbf{1.000}  &   0.872  &   0.724  &   \textbf{0.995}  &   0.971  &   0.830  &   0.853  &   0.973  &   0.893  &   \textbf{0.982}  &   0.742  &   \textbf{0.998}  &   0.983  &   0.962  &   0.958  &   0.874  &   0.926  &   0.956  &   \textbf{0.993}  &  7/21 \\
        \noalign{\vskip 1mm} 
        \hline

    \end{tabular}%
    }
    \caption{\emph{Endo-Depth} results in {Hamlyn} without scaling.  $\text{Abs}_{\text{rel}}$ and RMSE in [mm], $\text{RMSE}_{\text{log}}$ in [log mm]. We compare two losses that predict real scale: (Stereo) and its combination with monocular self-supervision (Mono + Stereo). \nth{1} column: metrics used for evaluation. \nth{2} column: losses used during training. Last column: number of times a method is the best for a given metric. Rest of the columns: results for the {Hamlyn} sequences.}
    \hspace{-15mm}
    \label{tab:endodepth_eval_scale_red}
\end{table*}

\begin{table}[t!]
    \small
    \centering
    \resizebox{\linewidth}{!} {%
    \begin{tabular}{ c  c  c  c  c  c  }
        \multicolumn{2}{c}{} & \multicolumn{4}{c}{\textbf{Sequence}} \\
        \noalign{\vskip 1mm} 
        \cline{3-6}
        \noalign{\vskip 1mm}   
        \multicolumn{1}{c}{} & \multicolumn{1}{c}{\textbf{Method}} & 1 & 4 & 19 & 20 \\
        \noalign{\vskip 1mm} 
        \hline
        \noalign{\vskip 1mm}  
        
        \cellcolor{Dandelion} & {LapDepth} \cite{song2021monocular} & {0.504}       & {0.432}       & {1.234}        & {0.847} \\
        \cellcolor{Dandelion} & IsoNRSfM \cite{parashar2017IsoNRSfM} &\underline{0.097}        &      0.048 & {\bf 0.062} &{\bf 0.064} \\
        \cellcolor{Dandelion} & \textbf{ours} Mono  &  0.293        & 0.054      & 0.235       & 0.134   \\ 
        \cellcolor{Dandelion} & \textbf{ours} Stereo   &   {\bf0.083}&\underline{0.024}&\underline{0.169}   &0.120  \\ 
        \cellcolor{Dandelion} \multirow{-5}{*}{Abs Rel}& \textbf{ours} Mono + Stereo   &   0.213  &{\bf0.023}&0.284& \underline{0.087}\\ 
        \noalign{\vskip 1mm} 
        \hline
        \noalign{\vskip 1mm}   
        
        \cellcolor{Dandelion} & {LapDepth} \cite{song2021monocular} & {29.132}       & {12.182}       & {75.260}        & {39.121} \\
        \cellcolor{Dandelion} & IsoNRSfM \cite{parashar2017IsoNRSfM} &\underline{2.563}        & 0.185      & {\bf 0.769} &{\bf 0.578}\\
        \cellcolor{Dandelion} & \textbf{ours} Mono  &  26.497      & 0.204      & 7.432       & 1.487   \\ 
        \cellcolor{Dandelion} & \textbf{ours} Stereo   &   {\bf1.379}&{\bf0.051}&\underline{5.275}&1.212  \\ 
        \cellcolor{Dandelion} \multirow{-5}{*}{Sq Rel}& \textbf{ours} Mono + Stereo   &   11.822 & \underline{0.052} & 11.768 & \underline{0.603} \\ 
        \noalign{\vskip 1mm} 
        \hline
        \noalign{\vskip 1mm} 
        
        \cellcolor{Dandelion} & {LapDepth} \cite{song2021monocular} & {20.710}       & {11.742}       & {26.742}        & {15.599} \\
        \cellcolor{Dandelion} & IsoNRSfM \cite{parashar2017IsoNRSfM} &\underline{19.815}       &2.698       &{\bf 6.478}  & \underline{6.420}\\
        \cellcolor{Dandelion} & \textbf{ours} Mono  &  34.382       & 2.937      & 18.640      & 8.354  \\ 
        \cellcolor{Dandelion} & \textbf{ours} Stereo   &   {\bf9.361}&\underline{1.432}&\underline{16.213}  &7.491  \\ 
        \cellcolor{Dandelion} \multirow{-5}{*}{RMSE}& \textbf{ours} Mono + Stereo   &   23.553 &{\bf1.428}  &  24.026&{\bf5.401} \\ 
        \noalign{\vskip 1mm} 
        \hline
        \noalign{\vskip 1mm} 
        
        \cellcolor{Dandelion} & {LapDepth \cite{song2021monocular}} & {0.417}       & {0.408}       & {0.703}        & {0.575} \\
        \cellcolor{Dandelion} & IsoNRSfM \cite{parashar2017IsoNRSfM} &{\underline{0.122}}  &{{0.063}} &{{\bf 0.095}}  &{{\bf 0.084}}\\
        \cellcolor{Dandelion} & \textbf{ours} Mono  &  0.310       & 0.068      & 0.264       & 0.168   \\ 
        \cellcolor{Dandelion} & \textbf{ours} Stereo   &   {\bf0.109}  & {\bf0.032} & \underline{0.228}  & 0.147 \\ 
        \cellcolor{Dandelion} \multirow{-5}{*}{RMSE\textsubscript{log}}& \textbf{ours} Mono + Stereo   &   0.251  & \bf0.032 & 0.328  & \underline{0.108} \\ 
        \noalign{\vskip 1mm} 
        \hline
        \noalign{\vskip 1mm} 
        
        \cellcolor{Orchid} & {LapDepth} \cite{song2021monocular} & {0.692}       & {0.668}       & {0.493}        & {0.649} \\
        \cellcolor{Orchid} & IsoNRSfM \cite{parashar2017IsoNRSfM} &{\bf 0.930}  &   0.997    &{\bf 0.960}  &{\bf 0.988}\\
        \cellcolor{Orchid} & \textbf{ours} Mono  & 0.741       & 0.997      & 0.595       & 0.799   \\ 
        \cellcolor{Orchid} & \textbf{ours} Stereo   &   \underline{0.926}  &{\bf1.000}&\underline{0.796}& 0.845 \\ 
        \cellcolor{Orchid} \multirow{-5}{*}{$\delta < 1.25^{1}$}& \textbf{ours} Mono + Stereo   &   0.762  &{\bf1.000}&0.689& \underline{0.959}\\ 
        \noalign{\vskip 1mm} 
        \hline
        \noalign{\vskip 1mm} 
        
        \cellcolor{Orchid} & {LapDepth} \cite{song2021monocular} & {0.917}       & {0.925}       & {0.762}        & {0.875} \\
        \cellcolor{Orchid} & IsoNRSfM \cite{parashar2017IsoNRSfM} &{\bf 0.998}  &{\bf1.000}  &{\bf 0.997}  &{\bf 1.000}\\
        \cellcolor{Orchid} & \textbf{ours} Mono  & 0.881       & {\bf1.000} & \underline{0.921}       & 0.988   \\ 
        \cellcolor{Orchid} & \textbf{ours} Stereo   &   \underline{0.988}  &{\bf1.000}& 0.910  & 0.995 \\ 
        \cellcolor{Orchid} \multirow{-5}{*}{$\delta < 1.25^2$}& \textbf{ours} Mono + Stereo   &   0.931  &{\bf1.000} & 0.832  & \underline{0.999}\\ 
        \noalign{\vskip 1mm} 
        \hline
        \noalign{\vskip 1mm} 
        
        \cellcolor{Orchid} & {LapDepth} \cite{song2021monocular} & {0.964}       & {0.967}       & {0.872}        & {0.930} \\
        \cellcolor{Orchid} & IsoNRSfM \cite{parashar2017IsoNRSfM} &{\bf 0.999}  &{\bf1.000}  &{\bf 1.000}  &{\bf1.000}\\
        \cellcolor{Orchid} & \textbf{ours} Mono  & 0.928       & {\bf1.000} & \underline{0.988}       & 0.999 \\ 
        \cellcolor{Orchid} & \textbf{ours} Stereo   &   \underline{0.998}  &{\bf1.000}& 0.964  &{\bf1.000} \\ 
        \cellcolor{Orchid} \multirow{-5}{*}{$\delta < 1.25^3$}& \textbf{ours} Mono + Stereo   &  0.965  &{\bf1.000} & 0.909  &{\bf1.000} \\ 
        \noalign{\vskip 1mm} 
        \hline
        \noalign{\vskip 1mm} 
        
    \end{tabular}%
    }
    \caption{{Comparison against baselines \cite{parashar2017IsoNRSfM} and \cite{song2021monocular} in {Hamlyn}.  $\text{Abs}_{\text{rel}}$ and RMSE in [mm], $\text{RMSE}_{\text{log}}$ in [log mm]. Best results are boldfaced, second best are underlined. }}
    \vspace{-1mm}
    \label{tab:endodepth_eval_red}
\end{table}

{Table \ref{tab:endodepth_eval_red} shows the comparison between \emph{Endo-Depth} and the baselines LapDepth~\cite{song2021monocular} and IsoNRSfM~\cite{parashar2017IsoNRSfM}. LapDepth is a state-of-the-art supervised depth network, that we trained in the synthetic dataset of \cite{rau2019implicit}, containing $16K$ images, using the authors' default configuration. {IsoNRSfM} is a multi-view method for which we chose the state-of-the-art implementation of \cite{rodriguez2020sd}. However, we re-ran IsoNRSfM and obtained even better results than \cite{rodriguez2020sd}. We use the per-image scaling as in~\cite{rodriguez2020sd}, $s_i^{\prime} = \text{median}\left(\nicefrac{D_i^{GT}}{D_i} \right)$. Although it is slightly different from the one in Table \ref{tab:endodepth_eval_red_without_scale}, we prefer to keep the metrics strictly as they were defined in each of our references. Notice that Table \ref{tab:endodepth_eval_red} reports IsoNRSfM results only in keyframes and only for the sparse point cloud, while \emph{Endo-Depth} {and LapDepth} report dense ones.} Note also that we are able to run \emph{Endo-Depth} in a larger set of {Hamlyn} sequences than IsoNRSfM (21 of them, see Tables \ref{tab:endodepth_eval_red_without_scale} and \ref{tab:endodepth_eval_scale_red}). We compare the {three} methods in {four} different sequences. In video \#1 the camera explores the abdominal wall from a distant point causing a small stereo parallax, hindering the evaluation. In video \#4, the camera is static and captures a beating heart. Video \#19, shows challenging organ deformations with low texture and tool intrusion. Lastly, video \#20, images a deforming intestine. Such deformations and tools intrusions challenge the assumptions made by the monocular self-supervision, but our results are still competitive. 
\emph{Endo-Depth} has three main advantages for deformable mapping versus IsoNRSfM. First, we recover the dense depth of the entire image without the limitation of finding multi-view correspondences. Second, as shown in Fig. \ref{fig:depths}, we can cope with discontinuities. And finally, we can do it with a single-shot, allowing us to initialize depth without a sliding window of images. This last point is crucial in intracorporeal sequences where the lack of texture compromises the matching. {Finally, observe how LapDepth performs significantly worse than \emph{Endo-Depth} even if it was trained in supervised data. The reason is the domain shift between the synthetic and real images, which is our motivation to use self-supervised training.}



The computation time of \emph{Endo-Depth} for one test image is $\sim$$15$ ms in a Nvidia RTX 2080Ti GPU and $\sim$$55$ ms in a AMD Ryzen 9 3900X CPU. {IsoNRSfM}, for the same image, including warps and alignment, takes $\sim 500$ ms in a Intel i7-7700K CPU. For a better visualization of \emph{Endo-Depth} results, the reader is referred to our supplementary videos, that contain depth predictions for five {Hamlyn} test videos with our best training setup (stereo loss), together with their backprojected 3D point clouds and the ground truth\footnote{The video is at 
\url{https://youtu.be/V3Be2W3iomI?t=0}}.

\subsection{Camera Tracking}
\label{sec:camtracking}
Fig. \ref{fig:odometry_comparative} and Fig. \ref{fig:odometry_comparative_frames} show qualitative and quantitative results of our photometric tracking versus other alternatives. Specifically, we compare it against the Open3D~\cite{Zhou2018} implementations of ICP (using point-to-point and point-to-plane distances) and pose tracking using photometric and hybrid photometric and geometric residuals (at optimal keyframe creation ratios). We also show different keyframe creation ratios in Fig. \ref{fig:odometry_comparative}, being the higher ones the more convenient due to fast camera dynamics in these sequences. The computation time per frame of our Python implementation is around $300$ ms in our GPU (Nvidia RTX 2080Ti) and $700$ ms in CPU (AMD Ryzen 9 3900X).

\subsection{Volumetric Fusion}

\begin{figure*}[ht!]
    \centering
    \includegraphics[width=\textwidth,keepaspectratio]{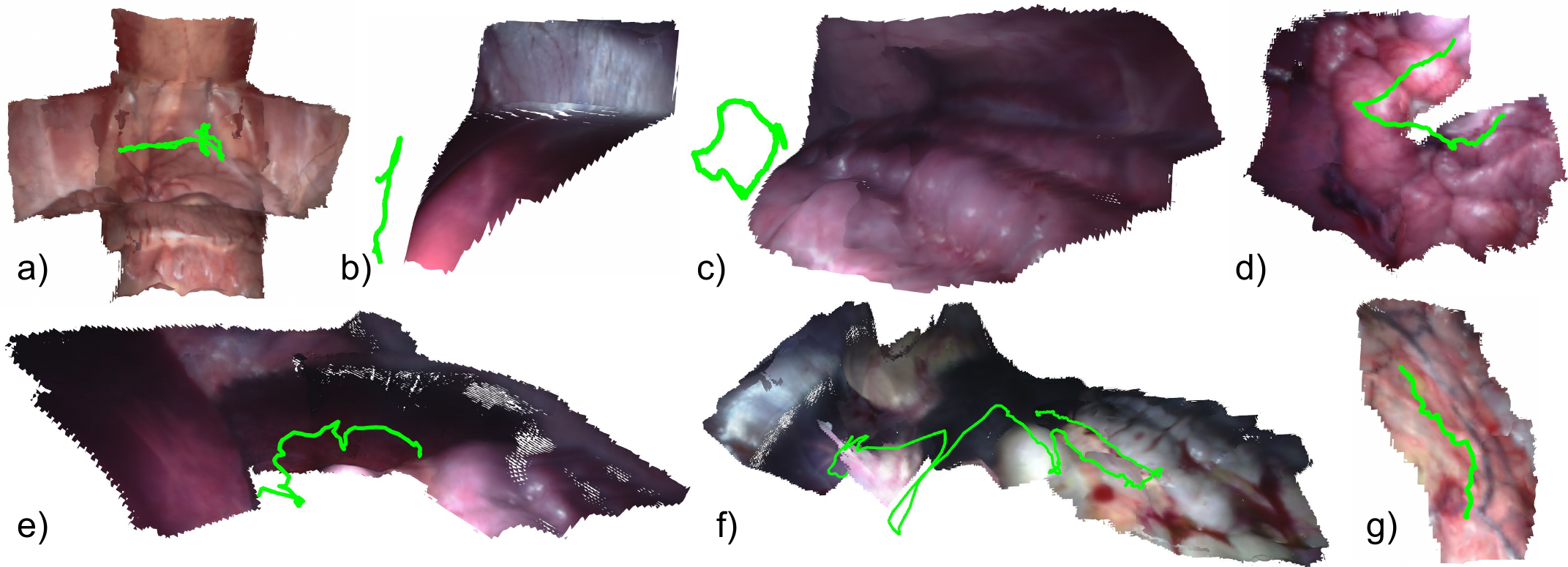}
    \captionof{figure}{\centering Maps after volumetric fusion of several {Hamlyn} videos: a) \#1, b) \#15, c) \#17. d) \#20, e) \#21, f) \#25 and g) \#27.}
    \label{fig:optimized_maps}
    \hspace{-15mm}
\end{figure*}

Fig. \ref{fig:optimized_maps} shows sample 3D reconstructions after fusing the predicted depth maps in a TSDF representation in 7 Hamlyn videos, chosen among those with larger camera translation. Although the ground truth for the scene and camera trajectory is not available, the accuracy of the reconstruction can be qualitatively assessed in the figure. Notice that we present a higher number of reconstructions and of larger extent than other approaches in the literature (e.g., \cite{song2018mis,lamarca2020defslam,rodriguez2020sd}), which again highlights the potential of depth networks, photometric \emph{pseudo-}RGBD tracking and volumetric fusion for reconstruction and motion estimation in monocular endoscopies. The Python implementation runs within reasonable time limits, the volumetric fusion of a $1200$-frames reconstruction taking $\sim$$7$ s in our CPU (AMD Ryzen 9 3900X). 

\section{Conclusions and Future Work}
\label{sec:9}

In this paper we have presented \emph{Endo-Depth-and-Motion}, in which we train and evaluate thoroughly state-of-the-art self-supervised depth learning in endoscopic videos, implement a robust photometric odometry and integrate their outputs with a volumetric fusion approach. Several conclusions can be extracted from our evaluation. Firstly, our self-supervised depth network \emph{Endo-Depth}, trained in the Hamlyn sequences, has a competitive performance even when compared against well-established multi-view baselines such as IsoNRSfM {and supervised networks as LapDepth}. This suggests that relevant future work could come out from the replacement of IsoNRSfM templates {and supervised depth learning} in SfM/SLAM pipelines for endoscopies. Importantly, such performance does not degrade when evaluated without per-image scaling, showing that the global scale of the scene can be effectively learned using stereo losses. Secondly, we implement a photometric dense tracking for estimating the camera motion with respect a \emph{pseudo-}RGBD keyframe. Finally, we fuse the predicted depth maps of the registered keyframes using a TSDF representation, showing again a {satisfactory} performance. {Our main line for future work is including deformation models in our scene representation.}


{
\balance
\bibliographystyle{IEEEtran}
\bibliography{biblio}
}








\end{document}